\documentclass[sigconf,nonacm]{acmart}
\AtBeginDocument{}
\setcopyright{none}
\acmDOI{}
\acmISBN{}
\acmConference{}{}{}
\usepackage{booktabs}
\usepackage{tabularx}
\usepackage{longtable}
\usepackage{array}
\usepackage{xcolor}
\usepackage{graphicx}
\usepackage{enumitem}

\newcommand{\para}[1]{\vskip 4pt\noindent\textbf{#1}\hskip .05in}

\newcolumntype{Y}{>{\raggedright\arraybackslash}X}
\newcommand{\tightcaption}[2]{\caption{\textbf{#1.} #2}}
\newcommand{\appendixnote}[1]{\noindent{\footnotesize #1}\par}

\newcommand{\arxivacceptancenote}{This work has been accepted for publication in the proceedings of the 19th ACM Workshop on Artificial Intelligence and Security (AISec 2026), co-located with ACM CCS 2026. The final version will be published in the ACM Digital Library.}
\makeatletter
\apptocmd{\@mkauthors}{%
  \global\setbox\mktitle@bx=\vbox{%
    \unvbox\mktitle@bx
    \hsize=\textwidth
    \normalfont\small\centering
    \arxivacceptancenote\par\bigskip
  }%
}{}{\PackageError{arxiv-notice}{Could not position the acceptance notice}{}}
\makeatother
\begin{document}
\title[What Breaks Local Watermarks?]{What Breaks Local Watermarks? A Robustness Benchmark for Local Invisible Image Watermarking}
\settopmatter{authorsperrow=3}

\author{Kai Yao}
\authornote{Both authors contributed equally to this research.}
\affiliation{
  \institution{University of Edinburgh}
  \city{Edinburgh}
  \country{UK}
}
\email{kai.yao@ed.ac.uk}

\author{Bence Szil{\'a}gyi}
\authornotemark[1]
\affiliation{
  \institution{Garandor}
  \city{Edinburgh}
  \country{UK}
}
\email{bence@garandor.com}

\author{Sebesty{\'e}n Kamp}
\affiliation[obeypunctuation=true]{
  \institution{Garandor}; \institution{University of Edinburgh}\\
  \city{Edinburgh}, \country{UK}
}
\email{seb@garandor.com}

\author{M{\'a}t{\'e} Po{\'o}r}
\affiliation{
  \institution{Garandor}
  \city{Edinburgh}
  \country{UK}
}
\email{mate.poor@garandor.com}

\author{M{\'a}t{\'e} Szilveszter}
\affiliation{
  \institution{Garandor}
  \city{Edinburgh}
  \country{UK}
}
\email{mate@garandor.com}

\author{Matyas K. Zsoldos}
\affiliation{
  \institution{Garandor}
  \city{Edinburgh}
  \country{UK}
}
\email{matyas@garandor.com}

\author{Marc Juarez}
\authornote{Corresponding author.}
\affiliation{
  \institution{University of Edinburgh}
  \city{Edinburgh}
  \country{UK}
}
\email{marc.juarez@ed.ac.uk}

\renewcommand{\shortauthors}{Yao et al.}

\begin{abstract}
Local image watermarking embeds an invisible signal into selected image regions rather than spreading it across the entire image, enabling payload recovery from specific objects or regions without perceptibly altering the image.
Existing studies evaluate the robustness of payload recovery and localization under image transformations, but they often focus on their own proposed method, resulting in narrow evaluations with inconsistent choices of transformations, datasets, and metrics.
These inconsistencies across studies limit direct comparisons across methods and muddle the overall picture of local watermark robustness.
To address this gap, we present the first systematic robustness benchmark for local watermarks across 55 image transformations, including (i) signal distortions, (ii) changes in image coordinate alignment, (iii) indirect local edits, and (iv) direct watermark edits.
The benchmark evaluates MaskWM, WAM, OmniGuard, TrustMark, and PixelSeal, all methods that either provide native localization or require minimal adaptation to support it.
Our results show that all evaluated methods are vulnerable to some transformation, with MaskWM standing out as offering the strongest payload recovery and localization, although it has the lowest image quality in the clean setting.
Synchronization further improves MaskWM's payload recovery under several geometric transformations, albeit at an additional cost to image quality.
A key finding is that local watermark robustness depends strongly on the nature of the transformation: signal distortions are often tolerated by the strongest methods, while geometric misalignment and generative local edits, such as inpainting and outpainting, can completely impair payload recovery.
We observe that payload recovery and localization are related but not interchangeable, and both strongly depend on the transformation's impact on the watermark region.

\end{abstract}




\maketitle

\section{Introduction}
\label{sec:introduction}
Generative image models have enabled new forms of image manipulation: objects can be easily inserted, erased, or restyled, and images increasingly combine content from multiple human and AI sources~\cite{rombach2022highresolution,yu2023inpaint}.
In response, local image watermarking has emerged as a leading approach for protecting intellectual property under such complex image manipulations~\cite{wam,maskwm}.
As opposed to {\it global} watermarking, which spreads a detectable signal across the entire image, local watermarking embeds the signal in specific regions, so that the watermark can still be detected even when the image has been cropped, outpainted, or spliced with other content.

Several methods to address this problem have been proposed in the academic literature.
However, the robustness of local watermarks remains difficult to compare, because existing studies tend to focus on the methods they propose and differ in their scheme configurations, choices of datasets, considered image transformations, localization targets, and evaluation metrics~\cite{wam,maskwm}.
This lack of a systematic evaluation makes it difficult to directly compare schemes and to quantify their relative robustness to manipulation.

To address this gap, we present a unified evaluation framework that measures robustness of five state-of-the-art localization-related watermarking methods to 55 image transformations across four datasets.
The suite includes four main categories of transformations, including conventional image distortions such as (i) signal distortions, which alter pixel values or frequency content while preserving image coordinates, and (ii) changes in alignment, which modify the image coordinate frame and canvas extent.
It also covers two types of local edits: (iii) indirect local edits and (iv) direct watermark edits, including splicing and AI-based inpainting and outpainting---transformations that are underrepresented in existing evaluations despite their relevance for local watermarking.

Besides robustness, these schemes aim to keep the watermark imperceptible and, in some cases, to segment the region where it was embedded.
Our evaluation thus covers three axes: payload recovery, i.e., whether the embedded message remains decodable; imperceptibility, approximated by image fidelity in the clean setting; and localization accuracy, measured as the Intersection over Union (IoU) between the predicted and ground-truth watermarked regions.
Each method is also evaluated with and without SyncSeal~\cite{syncseal}, an auxiliary {\it synchronization} technique that embeds an additional signal to correct geometric image transformations and restore the coordinate frame expected by the watermark decoder.

Our results show that no evaluated method is robust across all transformations.
In the clean setting, MaskWM achieves the strongest payload recovery and localization accuracy, but its watermarked images have the lowest fidelity.
Performance also varies with watermark-region area: smaller regions are associated with weaker payload recovery and, for MaskWM and WAM, worse localization accuracy.
Under transformations, for (i) signal distortion, only MaskWM remains robust across all five signal-distortion subcategories; for (ii) changes in alignment, we observe that geometric misalignment causes the sharpest reduction in payload recovery. SyncSeal partly restores the recovery lost to geometric misalignment, although its benefit varies widely across methods and transformations; for (iii) indirect local edits, the payload recovery can fail for generative edits even when there's no overlap between the edit and the watermarked region; and for (iv) direct watermark edits, replacing or regenerating the region where the watermark was embedded pushes recovery to random decoding for nearly all methods. Taken together, our results indicate that current local watermarking methods remain too brittle across realistic image manipulations for reliable production deployment.

Because the transformations we evaluate are non-adaptive, i.e., their parameters are not tuned against any individual scheme, our results have direct security consequences: a scheme's lack of robustness to a transformation in our benchmark implies a practical {\it black-box} attack that removes the watermark.
This means that the schemes are vulnerable to simple attacks, agnostic to the watermarking scheme and underlying generative models. Given the limited robustness we observe, these schemes are unlikely to provide provenance in realistic copyright scenarios, where there are adversaries with vested interests in removing the watermarks.

\smallskip
\noindent\textbf{Our contributions.}
\begin{itemize}[leftmargin=*]
    \item We introduce the first systematic local invisible image watermarking benchmark that evaluates payload recovery, localization accuracy, and imperceptibility, and separately assesses the effect of optional synchronization.
    \item Our benchmark evaluates five methods on four datasets under 55 non-adaptive transformations spanning four categories: (i) signal distortions, (ii) changes in alignment, (iii) indirect local edits, and (iv) direct watermark edits.
    \item We distinguish and analyze failures caused by signal distortion, misalignment, pixel replacement or regeneration, and diffusion-based reconstruction. We show that payload recovery and localization can fail independently, identify when synchronization helps or harms recovery, and highlight implications for watermark detection evasion.
\end{itemize}
The benchmark code will be available at \url{https://github.com/garandorinc/markbench}.

\section{Background and Related Work}
\label{sec:background}

In this section, we first introduce \emph{local watermarking}, a subcategory of image watermarking that embeds the payload message within a selected region to enable provenance of individual image objects. We then distinguish local watermarking from the closely related task of \emph{watermark-based tamper localization}. Finally, we review the literature on robustness evaluations of local watermarking.

\subsection{Local watermarking}
\label{subsec:local_watermarking}

Image watermarking embeds a hidden payload through changes intended to remain visually {\it imperceptible}~\cite{cox1997secure,hartung1999multimedia,cayre2005watermarking}. While early methods were based on signal-processing techniques, state-of-the-art approaches jointly train a neural network encoder--decoder pipeline that learns to embed and recover the payload from the image, subject to an imperceptibility constraint.

Traditional image watermarking methods have been {\it global}, meaning that the watermark signal was spread over the full image~\cite{synthid_image,pixelseal,trustmark}.
However, as generative models enable new forms of image manipulation and images increasingly combine content from multiple sources, there is a growing need to provide provenance of individual objects within an image, rather than of the entire image.
\emph{Local watermarking} has emerged as the leading technical approach to address these challenges.

Local watermarking embeds the watermark payload in specific regions and, during detection, not only decodes the payload but also identifies where these regions are. Representative methods include the Watermark Anything Model (WAM)~\cite{wam} and MaskWM~\cite{maskwm}. WAM reformulates watermark extraction as a segmentation problem: its extractor predicts a watermark-presence score and payload bits at each pixel, enabling pixel-level localization and payload recovery~\cite{wam}. MaskWM introduces {\it mask-guided} training into the standard encoder--decoder watermarking pipeline. By applying random image masks on regions of the images, it restricts the encoder to specific regions and teaches the decoder where the watermark signal is confined. Through a dedicated localization module, MaskWM can also identify watermarked regions before payload extraction~\cite{maskwm}.

A related but distinct direction of work is designed specifically for \emph{tamper localization}--estimating where an image has been edited or manipulated. Tamper localization methods also embed payloads into the image, but they aim to estimate a manipulation mask indicating where the protected image has been altered after payload embedding. For example, OmniGuard embeds two complementary signals: a payload for copyright protection and another payload for localization. Local edits can selectively corrupt the recovered localization payload, while the copyright payload is designed to remain recoverable~\cite{editguard,omniguard}.

\subsection{Robustness evaluations}

A substantial body of work has evaluated the robustness of conventional full-image watermarking against common signal-processing distortions and watermark removal attacks~\cite{voloshynovskiy2001attack,waves,eti_report}. More recent studies have further examined robustness against generative transformations and attacks enabled by modern image generation and editing models~\cite{zhao2024removable,diffusionbreak}. By contrast, robustness evaluation for local watermarking remains underexplored, with existing evidence derived primarily from method-specific evaluations introduced in WAM and MaskWM.

Sander et al.\ evaluate full-image watermark detection and payload recovery under geometric transformations, valuemetric distortions, splicing, and inpainting~\cite{wam}. For local watermarking, they construct images containing centered watermarked rectangles of varying sizes and evaluate watermark localization and payload recovery on these controlled composites, but do not apply additional distortions to the locally watermarked images for this evaluation. Thus, although WAM introduces local watermarking capabilities, it does not specifically evaluate their robustness under transformations or attacks.

Hu et al.\ go further by directly evaluating the robustness of local payload recovery and watermark-region localization~\cite{maskwm}. They construct a controlled local watermarking test set spanning twelve watermark-region size intervals from 1--5\% to 95--99\% of the full image. Across these region sizes, the authors evaluate and compare payload bit accuracy and localization performance under ten valuemetric distortions and three geometric transformations. However, this evaluation is built around a single mask compositing procedure and a limited set of conventional transformations. Therefore, it provides an important initial evaluation of local watermark robustness, but its results do not generalize to a diverse range of realistic transformations, including generative and local edits.

\section{Evaluation Framework}
\label{sec:benchmark-framework}
This section describes our evaluation framework in detail, including the watermarking methods, datasets, transformations and metrics that we used.

\subsection{Overview}

\begin{figure*}[htbp]
  \centering
  \includegraphics[width=\textwidth]{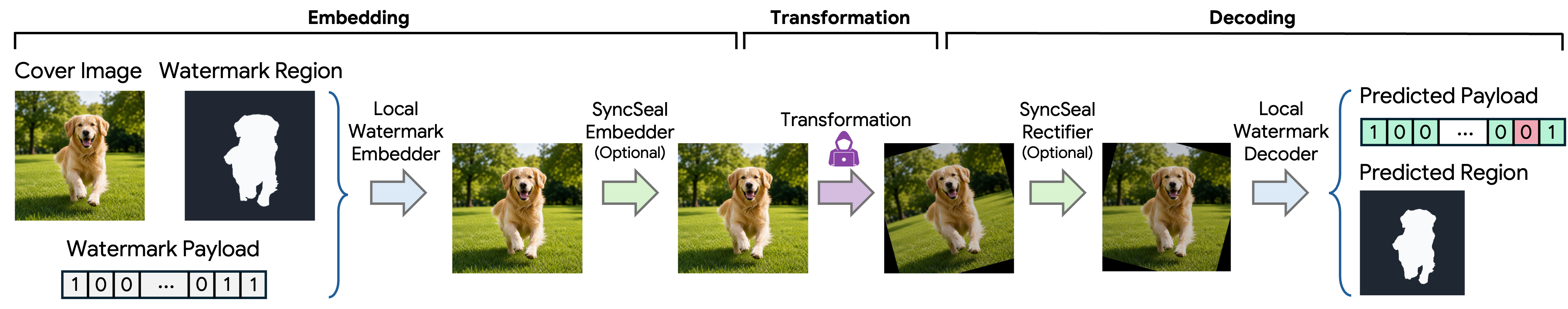}
  \Description{The evaluation proceeds through embedding, transformation, and decoding. Optional synchronization adds a signal before transformation and rectifies the image before decoding. Payload recovery, localization, and clean image fidelity are evaluated separately.}
  \tightcaption{Evaluation pipeline}{\emph{(i) Embedding.} Given a cover image, a watermark mask, and a binary payload, the watermarking method embeds the payload within the masked region.
  When SyncSeal is enabled, its embedder then adds a synchronization signal across the image.
  \emph{(ii) Transformation.} The resulting image undergoes one of the 55 benchmark transformations.
  \emph{(iii) Decoding.} When SyncSeal is enabled, its rectifier first attempts to restore the coordinate frame expected by the watermark decoder; otherwise, both SyncSeal operations are skipped.
  The watermark decoder then recovers the payload and, when supported, localizes the watermarked region.
  For runs with and without SyncSeal, we report payload recovery for all methods, pixel-level localization for MaskWM and WAM, and also image fidelity in the clean setting.}
  \label{fig:semantics}
\end{figure*}

Figure~\ref{fig:semantics} depicts one run of our evaluation pipeline, which is divided into three stages: {\bf embedding}, {\bf transformation}, and {\bf decoding}.

\begin{itemize}[leftmargin=*]
    \item In the {\bf embedding} stage, we draw an input image from our datasets and generate a watermark mask (Section~\ref{sec:datasets}). For each image and method, we sample a binary payload of the length supported by that method, drawing each bit uniformly from \(\{0,1\}\). The watermarking method then embeds the payload within the region defined by the watermark mask. The same payload is used across all image transformations, both with and without SyncSeal. When SyncSeal is enabled, the SyncSeal embedder adds a synchronization signal to the image after the payload embedding.

    \item In the {\bf transformation} stage, we apply an image transformation from our comprehensive taxonomy comprising 55 image manipulation functions (Section~\ref{sec:taxonomy}).

    \item The {\bf decoding} stage is {\it blind}: the decoder receives only a possibly transformed image, without access to the unwatermarked cover image. Then, if SyncSeal was used during embedding, it estimates the geometric transformation and attempts to map the image back to the coordinate frame expected by the decoder~\cite{syncseal}. The watermark decoder then performs payload recovery and, when supported, tries to identify the watermarked region.
\end{itemize}

\noindent
Finally, we assess the scheme's performance on an evaluation run along three dimensions: payload recovery, localization accuracy, and imperceptibility (Section~\ref{sec:metrics}).

\subsection{Localization and synchronization}\label{sec:methods}

We distinguish between two types of methods used in the watermarking pipeline: {\it localization} and {\it synchronization} methods.  

\para{Localization methods.} We evaluate five state-of-the-art local watermarking methods with different designs, each with a public repository and pretrained weights.
\textbf{MaskWM} and \textbf{WAM} are the two natively local watermarking methods evaluated in this work.
The other three methods (\textbf{OmniGuard}, \textbf{TrustMark}, and \textbf{PixelSeal}) were originally designed for global (i.e., full-image) watermarking, and we have adapted them for local watermarking by confining the embedding of the watermarks to a specific region.
We include \textbf{OmniGuard} because it is often compared with MaskWM and WAM for localization capability, though it was originally proposed for tamper localization, as described in Section~\ref{subsec:local_watermarking}.
\textbf{TrustMark} is relevant for the evaluation because its official implementation provides an optional watermark-region detector that returns a rectangular region for watermark localization capability~\cite{trustmark,trustmark_code}.
Finally, \textbf{PixelSeal} is a recent high-capacity global image and video watermarking method~\cite{pixelseal,videoseal_code}. We adapt it to a local watermarking setting to benchmark how state-of-the-art global watermarking techniques perform when only part of the embedded signal is utilized, compared to watermarking methods specifically designed for localization.
We summarize these embedding setups and decoder outputs in Table~\ref{tab:semantics}.
Appendix~\ref{app:local-adaptations} details the embedding and decoding procedures used to adapt OmniGuard, TrustMark, and PixelSeal to local watermarking.

\para{Synchronization methods.} Synchronization and watermark localization address different problems. Watermark localization refers to a decoder capability within the watermarking pipeline: identifying the image region that contains a local watermark. Synchronization, by contrast, is performed by a separate, optional module that estimates and reverses geometric transformations that may have been applied to the image before decoding. 

We evaluate SyncSeal as a representative standalone synchronization module that can be integrated with different watermarking methods~\cite{syncseal}. It embeds a learned synchronization signal across the image and uses the recovered signal to estimate the applied geometric transformation, allowing the image to be rectified before watermark decoding.

\begin{table*}[htbp]
\centering
\caption{Watermarking methods used in the evaluation. All methods are evaluated for payload recovery. For MaskWM and WAM, which return pixel-level maps of watermark presence, we report watermark-region localization using intersection-over-union (IoU). MaskWM and WAM are native local methods; OmniGuard, TrustMark, and PixelSeal are adapted reference methods originally designed for global (i.e., full-image) watermarking.}
\label{tab:semantics}
\small
\setlength{\tabcolsep}{4pt}
\begin{tabularx}{\textwidth}{@{}l >{\raggedright\arraybackslash}X c >{\raggedright\arraybackslash}X@{}}
\toprule
\textbf{Method} & \textbf{Embedding setup} & \textbf{Payload bits} & \textbf{Decoder output} \\
\midrule
MaskWM~\cite{maskwm} & Native embedding only within local regions & 64 & Payload + pixel-level map of watermark presence \\
WAM~\cite{wam} & Native embedding only within local regions & 32 & Payload + pixel-level map of watermark presence \\
OmniGuard~\cite{omniguard} & Global embedding confined to watermark region & 100 & Payload + tamper map$^{\dagger}$ \\
TrustMark~\cite{trustmark} & Global embedding confined to watermark region & 100 & Payload + rectangular region$^{*}$ \\
PixelSeal~\cite{pixelseal} & Global embedding confined to watermark region & 256 & Payload only \\
\addlinespace[0.25em]
\multicolumn{4}{@{}p{\dimexpr\textwidth-2\tabcolsep\relax}@{}}{\footnotesize $^{*}$ For TrustMark, the predicted watermark region is the rectangular output of its optional watermark-region detector.} \\
\multicolumn{4}{@{}p{\dimexpr\textwidth-2\tabcolsep\relax}@{}}{\footnotesize $^{\dagger}$ OmniGuard's tamper map estimates edited content rather than the watermark region.} \\
\bottomrule
\end{tabularx}
\end{table*}

\subsection{Image datasets and watermark masks}\label{sec:datasets}

For the evaluation, we selected four datasets chosen to cover different image distributions and mask sources.

\begin{itemize}[leftmargin=*]
\item  \textbf{COCO 2017 (COCO)}~\cite{coco} provides everyday scenes with instance segmentation masks, enabling object-aligned watermark regions in cluttered natural images.

\item \textbf{SA-1B}~\cite{kirillov2023segment} provides large-scale segmentation masks over diverse image content.

\item \textbf{DIV2K}~\cite{div2k} includes high-quality, high-resolution images originally curated for super-resolution, with fine texture and large spatial dimensions.

\item \textbf{MIRFLICKR-25K (MIRFLICKR)}~\cite{mirflickr} is a large in-the-wild photography dataset with diverse capture conditions and content.
\end{itemize}

\para{Watermark masks.}
For COCO and SA-1B, we use a segmentation mask provided with each image as the watermark region. DIV2K and MIRFLICKR do not provide segmentation masks, so we generate either a rectangular or an irregular region under the procedure described in Appendix~\ref{sec:empirical-details}. Although MaskWM and WAM support multiple watermark regions, our evaluation uses a single region per image to focus on robustness comparisons across schemes while limiting the additional complexity introduced by multi-region configurations.

\subsection{Transformation taxonomy}~\label{sec:taxonomy}

To comprehensively evaluate robustness of local watermarks, we include one clean setting as a baseline and 55 transformations.
These transformations are organized into a taxonomy of four categories depending on the nature of the underlying transformation mechanism as described below. The \emph{signal distortions} and \emph{changes in alignment} categories cover conventional image distortions that treat the image as a whole, without targeting any local regions. In contrast, we also evaluate two categories of local edits: \emph{indirect local edits} and \emph{direct watermark edits}. These four categories are mutually exclusive, with no transformation belonging to more than one category in our taxonomy. Table~\ref{tab:attack-taxonomy} in Appendix~\ref{sec:additional-evaluation-details} lists every transformation by category.
Appendix Table~\ref{tab:attack-implementation-summary} specifies the transformation operations and parameter settings; Appendix~\ref{app:transformation-parameters} explains their sampling.

\begin{itemize}[leftmargin=*]
\item  \textbf{Signal distortions} are transformations that preserve image coordinates but alter pixel values or frequency content, such as brightness adjustment, blurring and noising, JPEG compression, and perturbations to coefficients in the frequency domain.
\item  \textbf{Changes in alignment} cover transformations that change the image coordinate frame, such as cropping, rotation, flipping, and perspective warping, so the embedded watermark may no longer align with the coordinates expected by the decoder. To evaluate whether synchronization techniques can restore alignment, we evaluate each transformation with and without SyncSeal. This category also includes \emph{outpainting (expanded canvas)}, which resizes and places the watermarked image within a larger canvas before generating the surrounding content, thereby changing the scale and position of the original image content.
\item \textbf{Indirect local edits} include transformations that do not deliberately target the watermark region itself. \emph{Blackout} fills a randomly sampled region with black pixels, \emph{splicing} replaces pixels in a sampled region with content from another image in the same dataset, and \emph{inpainting} uses a Stable Diffusion inpainting pipeline to regenerate the selected region from the surrounding image context~\cite{rombach2022highresolution}. For these three transformations, the edit mask is sampled without reference to the watermark mask, so the two masks overlap only by chance. Because the diffusion pipeline may also alter pixels outside the edit mask, the \emph{inpainting (independent region, outside restored)} control run copies the watermarked input back outside that mask after generation. \emph{Outpainting (fixed canvas)} keeps the image size unchanged and uses the complement of the watermark region as the generation mask to outpaint.  Appendix Figure~\ref{fig:protocol-examples} illustrates the effect of inpainting and outpainting protocols for generative edits.

\item \textbf{Direct watermark edits} use the watermark region itself as the edit mask and replacement target, applying the same inpainting or splicing operations directly to this region. These conditions test whether any recoverable watermark signal remains after the pixels intended to carry the payload are replaced or regenerated.
\end{itemize}

\noindent
Together, these transformations capture realistic image manipulation workflows that alter or regenerate image content.

\subsection{Evaluation metrics}\label{sec:metrics}

We evaluate three main dimensions of local watermarking performance: (i) payload recovery, (ii) localization accuracy, and (iii) imperceptibility. Each dimension is characterized by specific metrics, as described below in detail.

\para{Payload recovery.} Payload recovery is quantified with two complementary metrics: {\it bit accuracy} and {\it evidence score}.
Bit accuracy is the fraction of payload bits decoded correctly, commonly used in the literature for assessing watermark recovery.
However, the payload lengths supported by the schemes range from 32 bits for WAM to 256 bits for PixelSeal, so the same bit accuracy does not provide the same evidence of payload recovery against random decoding for every method.
Hence, we report bit accuracy for consistency with previous evaluations but also report the evidence score, a measure that \emph{takes into account the payload length}, enabling a fairer comparison across methods with different payload lengths.
This evidence score is defined as the observed recovery against random decoding, under which each decoded bit matches the corresponding embedded bit independently with probability \(0.5\)~\cite{fernandez2024videosealopenefficient}.
For \(k\) correctly decoded bits in an \(n\)-bit payload, the probability that random decoding would produce at least \(k\) correct bits is \(p_{\mathrm{random}}=\Pr[\mathrm{Binomial}(n,0.5)\ge k]\). We report \(-\log_{10}(p_{\mathrm{random}})\), so larger values indicate stronger evidence of successful payload recovery.
A score of \(3\) corresponds to \(p_{\mathrm{random}}=10^{-3}\). Equivalently, if recovery were declared whenever the score is at least \(3\), the theoretical false positive rate for an individual decoding attempt would be at most \(0.1\%\), an operating point used in prior watermarking studies~\cite{waves}. We therefore use \(3\) as a qualitative reference level.

\para{Localization accuracy.} Watermark localization accuracy is measured by thresholding the predicted map at 0.5 to obtain a predicted watermark-region mask, consistent with MaskWM and WAM's reference implementations. then computing intersection-over-union (IoU) with the ground-truth watermark-region mask after applying the transformation's coordinate mapping.
The localization comparison is restricted to MaskWM and WAM because only these two methods return pixel-level maps of watermark presence, making them comparable to each other. In contrast, OmniGuard produces a tamper map indicating the region of manipulation instead of the region of watermark; TrustMark's detector returns a rectangular region and not a pixel-level map, creating a shape mismatch; and PixelSeal does not provide spatial output. Hence, the localization accuracy of these three methods cannot be directly compared to MaskWM and WAM.

\para{Imperceptibility.} We quantify the imperceptibility of the scheme by image fidelity in the clean setting, measured between the original cover image and the watermarked image, using metrics including PSNR, SSIM~\cite{wang2004ssim}, and LPIPS~\cite{zhang2018lpips}, which respectively capture pixel-level distortion, structural similarity, and perceptual similarity based on deep features. We evaluate only in the clean setting to isolate the visual distortion introduced by watermark embedding itself. After a transformation, fidelity metrics would conflate this distortion with changes caused by the transformation.

\section{Results}

In this section, we compare the watermarking methods along the three evaluation axes: payload recovery (Figure~\ref{fig:headline_payload_recovery}), watermark-region localization (Figure~\ref{fig:headline_loc_iou}), and image fidelity in the clean setting (Table~\ref{tab:clean-fidelity}).
We then further examine in detail how local watermark performance varies with watermark-region area (Figure~\ref{fig:watermark-area-vs-perf}), and is affected by generative local edits (Figure~\ref{fig:local-edit-log10p}) and SyncSeal synchronization (Figure~\ref{fig:sync-align-subattacks}).

For each method, we evaluate a fixed set of 200 images per dataset (800 in total). Unless otherwise stated, we report results without SyncSeal, and average values first within each dataset and then take the statistics across the four datasets. SyncSeal results and comparisons are labeled explicitly. More details on empirical evaluation are deferred to Appendix~\ref{sec:empirical-details}.

\subsection{Payload recovery}
\label{subsec:payload_recovery}
\begin{figure*}[htbp]
  \centering
  \includegraphics[width=0.9\textwidth]{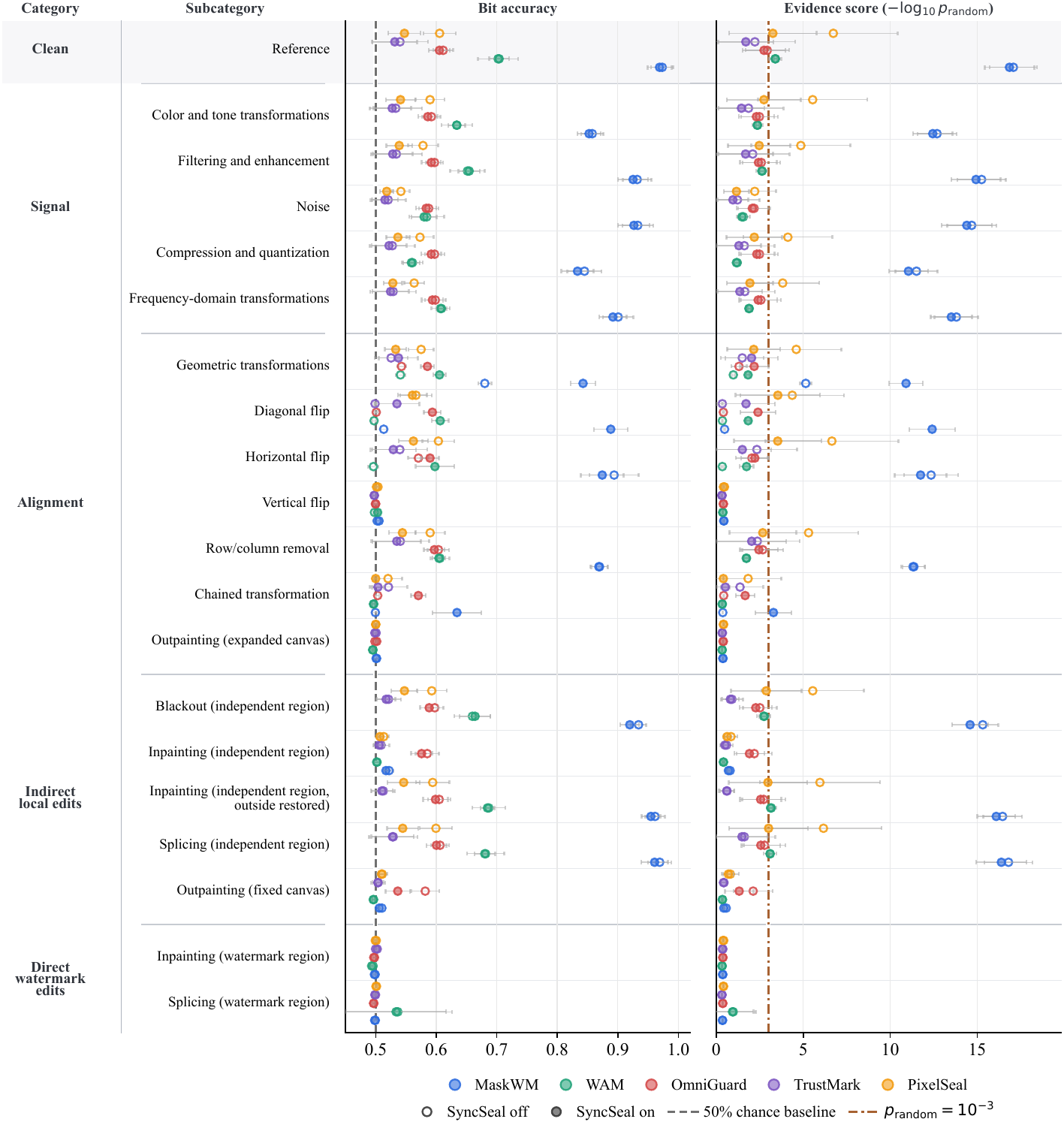}
  \Description{Payload recovery for five watermarking methods across the transformation taxonomy, with separate panels for bit accuracy and payload evidence and comparisons with and without synchronization.}
  \tightcaption{Payload recovery over the transformation taxonomy}{Rows follow the transformation taxonomy in Appendix Table~\ref{tab:attack-taxonomy}, with categories and subcategories shown at left.
  The left panel reports bit accuracy; the right panel reports the payload-length-aware evidence score, $-\log_{10}(p_{\mathrm{random}})$, where larger values indicate stronger evidence against random decoding after accounting for payload length.
  Open markers denote evaluation without SyncSeal and filled markers with SyncSeal; horizontal bars show the sample standard deviation across the four dataset means.
  The gray dashed line appears only in the bit-accuracy panel and marks the 50\% chance baseline; the brown dash-dot line in the evidence-score panel marks the threshold \(p_{\mathrm{random}}=10^{-3}\).}
  \label{fig:headline_payload_recovery}
\end{figure*}

Figure~\ref{fig:headline_payload_recovery} presents payload recovery across the transformations in terms of both bit accuracy and evidence score (see Table~\ref{tab:attack-taxonomy} in the appendix for details).
These results represent the broadest benchmark of our evaluation as it covers payload recovery for all five methods under every evaluation condition.

\para{Payload recovery in the clean setting.} The clean setting establishes how well each method recovers its payload before any transformation is applied. Our results show that payload recovery in the clean setting varies substantially across methods (first row of Figure~\ref{fig:headline_payload_recovery}).
MaskWM, one of the native local methods, gives the strongest recovery, with bit accuracy \(0.973\pm0.018\) and an evidence score of \(17.08\pm1.37\), though it has the lowest image fidelity in the clean setting among the evaluated methods (Section~\ref{sec:imperceptibility}).
WAM is also a native local method, but its clean-setting recovery is much weaker, with bit accuracy of \(0.702\pm0.033\) and an evidence score of \(3.39\pm0.35\).

One design difference between the two native local methods is how they use masks during training and decoding. WAM uses masks mainly for mask-based augmentation and pixel-level supervision, while MaskWM instead uses masks to guide local extraction and embedding directly, which may be one of the reasons why MaskWM offers better localization.

Notably, strong clean-setting recovery is not limited to the native local methods: PixelSeal reaches an evidence score of \(6.73\). Although PixelSeal is a global watermarking scheme adapted for localization, its evidence score even exceeds that of WAM in the clean setting. This result shows that a method need not be designed specifically for local embedding to provide strong evidence of local payload recovery.

\para{Payload recovery under transformations.} The clean-setting performance on payload recovery provides a baseline against which we assess the impact of image transformations on local watermarking methods. Overall, we find that payload recovery varies substantially across transformation categories and methods, and we provide a breakdown of the results by category below.

\para{(I) Signal distortions.} Signal distortions alter pixel values or frequency content without changing the image coordinate frame or replacing any local region. They therefore provide the most direct evaluation of the watermark signal itself: the payload region remains aligned with the coordinates expected by the decoder and is not directly replaced or regenerated.
Even in this setting, robustness remains strongly method-dependent. MaskWM's mean evidence score remains well above the threshold \(p_{\mathrm{random}}=10^{-3}\) in all five signal subcategories, with scores of \(11.51\)--\(15.25\). PixelSeal's mean score exceeds the threshold in four of five subcategories, ranging from \(3.82\) to \(5.54\), but falls below it under noise (\(2.21\)). The mean scores for WAM, OmniGuard, and TrustMark remain below the threshold in all five subcategories. Thus, only MaskWM shows consistently strong evidence of payload recovery across all five signal subcategories.

\para{(II) Changes in alignment.} Unlike signal distortions, misalignment disrupts the geometric correspondence between the pixels that carry the payload and those observed by the decoder. Its effect varies sharply across individual transformations.
For MaskWM without SyncSeal, the mean evidence score remains high under horizontal flip (\(12.35\)) and row/column removal (\(11.31\)), but falls below \(0.5\) under diagonal flip, vertical flip, chained transformation, and outpainting (expanded canvas).
SyncSeal improves recovery in some cases: MaskWM's mean score rises from \(5.14\) to \(10.91\) for geometric transformations and from \(0.48\) to \(12.41\) for diagonal flip, but remains below \(0.5\) for vertical flip and outpainting (expanded canvas).
Section~\ref{sec:syncseal-results} examines this effect in detail.

\para{(III) Indirect local edits.} Indirect local edits can be further distinguished by whether they modify only pixels within the edit region or also affect pixels outside it. This distinction splits indirect local edits into localized pixel replacement (e.g., blackout, splicing) and generative editing (e.g., inpainting, outpainting) which may induce whole-image reconstruction.
MaskWM and PixelSeal retain mean evidence scores above the threshold under blackout (independent region) and splicing (independent region).
Under the same setting, inpainting (independent region) and outpainting (fixed canvas) instead leave every watermarking method below the threshold.
However, under inpainting (independent region, outside restored), where the outside of the edit region is restored with the exact pixels before transformation, the mean score returns above the threshold for MaskWM, PixelSeal, and WAM. This control indicates that changes outside the edit mask induced by whole-image reconstruction contribute substantially to the loss of recovery. We analyze this further in detail in Section~\ref{sec:generative_local_edits}.

\para{(IV) Direct watermark edits.} For direct watermark edits, the distinction between localized replacement and broader reconstruction matters less because they both directly replace pixels in the watermarked region.
Our results confirm this: direct inpainting and splicing both leave every method with a mean evidence score below \(1\) and drive bit accuracy toward random decoding.

\subsection{Localization accuracy}
\begin{figure}[htbp]
  \centering
  \includegraphics[width=\columnwidth]{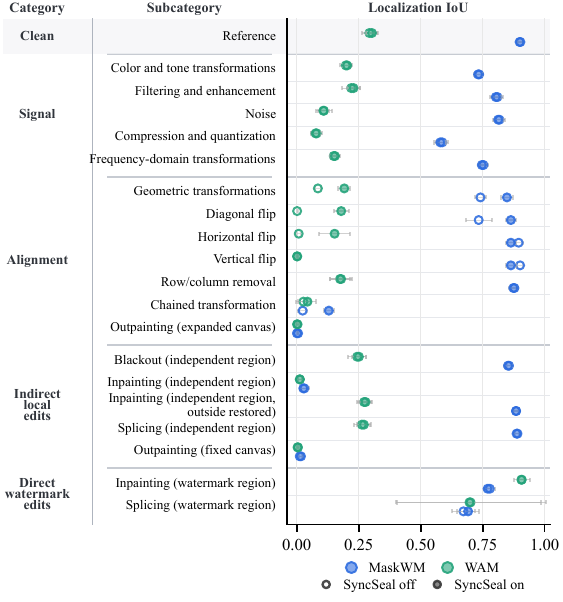}
  \Description{Watermark localization intersection over union for MaskWM and WAM across transformation categories, compared with and without synchronization.}

  \tightcaption{Localization accuracy over the transformation taxonomy}{Rows follow the transformation taxonomy in Appendix Table~\ref{tab:attack-taxonomy}.
  The plot reports localization IoU for MaskWM and WAM, the methods that return directly comparable pixel-level maps of watermark presence.
  Open markers denote SyncSeal off and filled markers denote SyncSeal on; horizontal bars show the sample standard deviation across the four dataset means.}
  \label{fig:headline_loc_iou}
\end{figure}

Besides payload recovery, another important indicator of local watermarking performance is the method's ability to localize the watermark region.
Figure~\ref{fig:headline_loc_iou} shows localization accuracy, measured by localization IoU, for MaskWM and WAM, the two methods that return directly comparable pixel-level maps of watermark presence. In the clean setting, MaskWM achieves a localization IoU of \(0.902\), substantially higher than WAM's \(0.293\), indicating stronger localization performance. Across transformations, localization degrades most sharply under \emph{changes in alignment}, which disrupt the spatial correspondence between the predicted watermark map and the original watermark region. By restoring this correspondence, SyncSeal raises MaskWM's IoU from \(0.741\) to \(0.848\) and WAM's from \(0.085\) to \(0.192\), showing that synchronization can partially recover localization performance under misalignment.

Some transformations separate localization from payload recovery, where the watermark region can be localized but payload recovery fails.
For example, under vertical flip, MaskWM's localization IoU remains high, at \(0.901\) without SyncSeal and \(0.864\) with SyncSeal, while the corresponding bit accuracies remain at random decoding (\(0.502\) and \(0.505\)).
Direct watermark inpainting and splicing show a similar distinction between localization and payload recovery.
Under inpainting, MaskWM and WAM retain localization IoUs of \(0.779\) and \(0.906\), respectively; under splicing, the corresponding IoUs are \(0.673\) and \(0.701\), demonstrating the capability to localize the watermark region even though both methods' bit accuracies remain at the random-decoding level for both inpainting and splicing.

These results show that localization and payload decoding can respond differently to the transformed signal under certain image transformations: a pixel-level map of watermark presence can remain informative even when bit accuracy is near random decoding. Payload recovery and localization therefore capture distinct failure modes and should be treated separately for local watermarking methods.

\subsection{Imperceptibility}
\label{sec:imperceptibility}
Payload recovery and localization characterize a local watermarking scheme's ability to recover the watermark and identify the region in which it is embedded. However, imperceptibility must be also considered, as it is a practical constraint in most local watermarking use cases.
Here we report image fidelity in the clean setting (Table~\ref{tab:clean-fidelity}) using multiple metrics as defined in Section~\ref{sec:metrics}. 

In the clean setting, MaskWM offers the strongest payload recovery and localization accuracy, but it also introduces the largest embedding distortion. Its mean clean-setting PSNR is \(49.48\,\mathrm{dB}\), significantly lower than the corresponding values for WAM (\(62.55\)), PixelSeal (\(56.80\)), TrustMark (\(54.09\)), and even OmniGuard (\(51.36\)).
MaskWM also has the highest mean LPIPS (\(0.006\), compared with values that are approximately \(0.000\) for the others), indicating lower image fidelity under this metric.
Mean SSIM remains uniformly high across methods, ranging from \(0.998\) to \(1.000\).

Importantly, the fidelity cost goes up when SyncSeal is enabled because SyncSeal adds a full-image synchronization signal. With SyncSeal enabled, mean PSNR remains above \(42\,\mathrm{dB}\) for all evaluated methods. Relative to the corresponding values without SyncSeal, mean PSNR drops by \(7.17\,\mathrm{dB}\) for MaskWM, \(18.99\,\mathrm{dB}\) for WAM, \(8.63\,\mathrm{dB}\) for OmniGuard, \(11.11\,\mathrm{dB}\) for TrustMark, and \(13.45\,\mathrm{dB}\) for PixelSeal. These results indicate that synchronization techniques, such as SyncSeal, should be applied carefully, weighing their gains in robustness against the associated image fidelity loss.

\begin{table}[htbp]
\centering
\caption{Image fidelity in the clean condition, used here as a proxy for imperceptibility. Rows report results with SyncSeal off and on; superscripts give the sample standard deviation across the four dataset means. Higher PSNR and SSIM and lower LPIPS indicate better image fidelity.}
\label{tab:clean-fidelity}
\scriptsize
\setlength{\tabcolsep}{2.0pt}
\begin{tabular*}{\columnwidth}{@{\extracolsep{\fill}}l c c c c c c}
\toprule
\textbf{Method} & \multicolumn{2}{c}{\textbf{PSNR (dB)}} & \multicolumn{2}{c}{\textbf{SSIM}} & \multicolumn{2}{c}{\textbf{LPIPS}} \\
\cmidrule(lr){2-3}\cmidrule(lr){4-5}\cmidrule(lr){6-7}
 & \shortstack{\textbf{SyncSeal}\\\textbf{off}} & \shortstack{\textbf{SyncSeal}\\\textbf{on}} & \shortstack{\textbf{SyncSeal}\\\textbf{off}} & \shortstack{\textbf{SyncSeal}\\\textbf{on}} & \shortstack{\textbf{SyncSeal}\\\textbf{off}} & \shortstack{\textbf{SyncSeal}\\\textbf{on}} \\
\midrule
MaskWM & 49.481{\textcolor{black!55}{\tiny$^{\pm 1.082}$}} & 42.309{\textcolor{black!55}{\tiny$^{\pm 0.617}$}} & 0.998{\textcolor{black!55}{\tiny$^{\pm 0.001}$}} & 0.990{\textcolor{black!55}{\tiny$^{\pm 0.004}$}} & 0.006{\textcolor{black!55}{\tiny$^{\pm 0.001}$}} & 0.009{\textcolor{black!55}{\tiny$^{\pm 0.002}$}} \\
WAM & 62.549{\textcolor{black!55}{\tiny$^{\pm 1.185}$}} & 43.562{\textcolor{black!55}{\tiny$^{\pm 0.824}$}} & 1.000{\textcolor{black!55}{\tiny$^{\pm 0.000}$}} & 0.992{\textcolor{black!55}{\tiny$^{\pm 0.003}$}} & 0.000{\textcolor{black!55}{\tiny$^{\pm 0.000}$}} & 0.003{\textcolor{black!55}{\tiny$^{\pm 0.001}$}} \\
PixelSeal & 56.800{\textcolor{black!55}{\tiny$^{\pm 1.243}$}} & 43.353{\textcolor{black!55}{\tiny$^{\pm 0.811}$}} & 0.999{\textcolor{black!55}{\tiny$^{\pm 0.000}$}} & 0.992{\textcolor{black!55}{\tiny$^{\pm 0.003}$}} & 0.000{\textcolor{black!55}{\tiny$^{\pm 0.000}$}} & 0.003{\textcolor{black!55}{\tiny$^{\pm 0.001}$}} \\
TrustMark & 54.092{\textcolor{black!55}{\tiny$^{\pm 2.075}$}} & 42.978{\textcolor{black!55}{\tiny$^{\pm 0.857}$}} & 0.999{\textcolor{black!55}{\tiny$^{\pm 0.001}$}} & 0.991{\textcolor{black!55}{\tiny$^{\pm 0.003}$}} & 0.000{\textcolor{black!55}{\tiny$^{\pm 0.000}$}} & 0.003{\textcolor{black!55}{\tiny$^{\pm 0.000}$}} \\
OmniGuard & 51.361{\textcolor{black!55}{\tiny$^{\pm 0.837}$}} & 42.730{\textcolor{black!55}{\tiny$^{\pm 0.841}$}} & 0.998{\textcolor{black!55}{\tiny$^{\pm 0.001}$}} & 0.991{\textcolor{black!55}{\tiny$^{\pm 0.003}$}} & 0.000{\textcolor{black!55}{\tiny$^{\pm 0.000}$}} & 0.003{\textcolor{black!55}{\tiny$^{\pm 0.001}$}} \\
\bottomrule
\end{tabular*}
\end{table}

\subsection{Performance by watermark-region area}

Watermarking method design is not the only source of variation in these performance outcomes. Because local embedding confines the watermark signal to a selected region, performance may also vary with the amount of image area available for embedding. We report how this variation unfolds below.

We use COCO and SA-1B for this analysis because their watermark regions are derived from native object-segmentation masks, whereas DIV2K and MIRFLICKR use procedurally randomly sampled masks; all four distributions are shown in Figure~\ref{fig:mask-area-dist} in the appendix.
The average watermark-region area is small across datasets: \(15.22\%\) for COCO, \(12.50\%\) for SA-1B, and \(8.80\%\) for DIV2K and MIRFLICKR.
Much of the evaluation therefore focuses on small regions, which both stress-tests the evaluated watermarking methods and reflects practical settings in which the provenance of small objects within an image must be tracked.

\begin{figure}[htbp]
  \centering
  \includegraphics[width=\columnwidth]{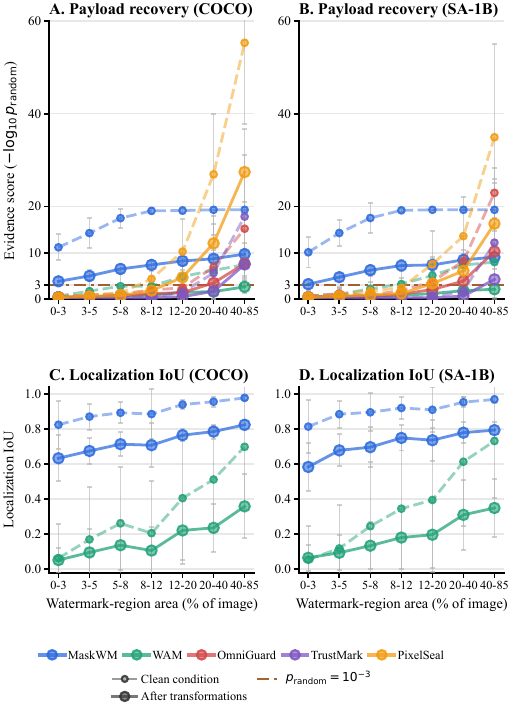}
  \Description{Payload recovery and watermark localization as a function of the fraction of image area occupied by the watermark region, shown for COCO and SA-1B.}
  \tightcaption{Larger watermark regions are associated with stronger payload recovery and watermark-region localization}{The four panels group COCO and SA-1B examples by the fraction of image area covered by the watermark region, evaluated without SyncSeal.
  Panels A and B report the evidence score.
  Panels C and D report localization IoU for MaskWM and WAM.
  Dashed curves show clean-setting values.
  For the solid curves, bit accuracy and localization IoU are first averaged across all transformations for each image, and the evidence score is computed from the resulting mean bit accuracy.
  The brown dash-dot line in the payload-recovery panels marks \(p_{\mathrm{random}}=10^{-3}\).
  Vertical bars show the sample standard deviation across image-level values within each area bin.}
  \label{fig:watermark-area-vs-perf}
\end{figure}

Across COCO and SA-1B, larger watermark regions are associated with higher payload-recovery evidence scores (Figure~\ref{fig:watermark-area-vs-perf}).
On COCO, MaskWM's post-transformation evidence score rises from \(3.84\) in the \(0\)--\(3\%\) bin to \(9.71\) in the \(40\)--\(85\%\) bin.
PixelSeal shows an even larger observed increase, from \(0.36\) to \(27.42\).
SA-1B shows the same trend.
One reasonable explanation is that a larger region provides the embedder with more pixels over which to distribute the payload and the decoder with more potential redundancy.

Localization accuracy follows a similar pattern to payload recovery across varying region areas.
MaskWM's post-transformation IoU increases from \(0.634\) to \(0.825\) on COCO with increased watermark-region area.
WAM increases from \(0.052\) to \(0.359\) over the same range, but remains below MaskWM.
WAM also degrades more from the clean setting, indicating that its localization output is less stable under transformations even for larger regions.

\subsection{A deeper analysis of generative edits}
\label{sec:generative_local_edits}
One important aspect of our benchmark is its evaluation of the effect of generative edits, a type of manipulation that is underrepresented in prior work. We therefore provide a deeper analysis here, complementing the results presented in Section~\ref{subsec:payload_recovery}, to explore which factors contribute to the degradation of payload recovery under inpainting transformations.

We first analyze whether this degradation is explained by the spatial relationship between the edit and watermark regions. We then examine whether restoring the changes outside the edit region recovers payload. Finally, we investigate whether characteristics of the watermark region, particularly its edge density, affect payload recovery under such edits.

\begin{figure*}[htbp]
  \centering
  \includegraphics[width=\textwidth]{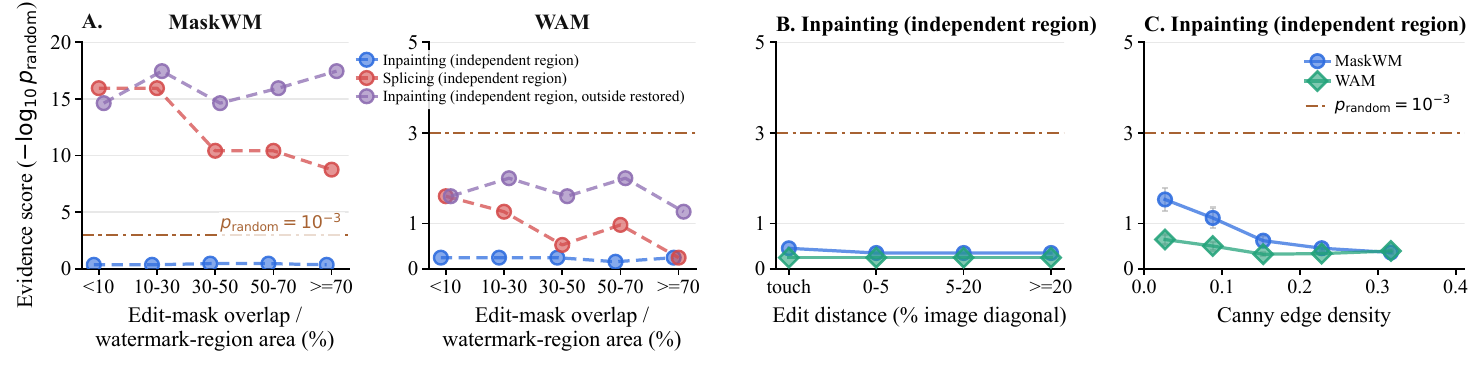}
  \Description{Payload recovery for MaskWM and WAM under inpainting and splicing as watermark overlap varies, followed by analyses of inpainting recovery against distance and edge density.}
  \tightcaption{Payload recovery under local edits}{The panels examine three factors in local edits: edit-mask overlap, distance from the watermark region, and edge density within preserved watermark-region pixels.
  Panel A compares inpainting (independent region), splicing (independent region), and the inpainting (independent region, outside restored) control for MaskWM (left) and WAM (right).
  Panel B shows that increasing the edit-to-watermark distance for inpainting (independent region) does not restore recovery.
  Panel C divides inpainting (independent region) examples with less than \(5\%\) edit-mask overlap into quintiles by Canny edge density over watermark-region pixels outside the edit mask.
  Markers show the mean edge density and evidence score in each quintile, and vertical bars show standard errors of the evidence-score means across examples within that quintile.
  MaskWM has higher evidence scores when those preserved pixels have lower Canny edge density, while WAM's evidence scores remain close to zero.
  Brown dash-dot lines mark the evidence-score threshold \(p_{\mathrm{random}}=10^{-3}\).}
  \label{fig:local-edit-log10p}
\end{figure*}

\para{Recovery does not improve with less overlap or greater distance.}
In Figure~\ref{fig:local-edit-log10p}, we first test whether overlap or distance between the edit region and the watermark region explains the poor recovery under inpainting in an independently sampled edit region.
If degradation were mainly caused by overlap with the watermark region, recovery should improve when the edit mask barely overlaps the watermark.
However, the results do not show improvement with less overlap: for MaskWM, the evidence score remains close to zero across the five overlap bins from \(<10\%\) to \(\geq70\%\) (Panel A, left, blue line); and WAM follows the same pattern (Panel A, right, blue line).

Increasing the distance between the edit region and the watermark region also does not restore recovery.
For example, MaskWM's evidence score is \(0.451\) when the edit touches the watermark region and \(0.346\) across the non-touching distance bins (Panel B).

\para{Restoring pixels outside the edit mask restores payload recovery.} Since overlap and distance do not explain the recovery loss, we next compare splicing, inpainting, and the outside-restored control to isolate the changes that the generative pipeline induces outside the edit mask.
Splicing an independently sampled edit region replaces only the edit region with external content and leaves surrounding pixels unchanged. As a result, MaskWM's splicing evidence score remains high across the overlap bins (\(15.95\), \(15.95\), \(10.42\), \(10.42\), \(8.75\)).
Inpainting, however, passes the entire image through the reconstruction pipeline, so pixels outside the edit mask change even when the image appears visually preserved. This drives inpainting's evidence score to near zero.

The outside-restored control tests whether changes outside the edit mask contribute to the loss of payload recovery: it restores all pixels outside the edit mask to their original values in the watermarked image. The results show that restoring these outside pixels recovers performance: MaskWM's evidence score remains high across all overlap bins (\(14.62\)--\(17.45\)), and WAM shows the same pattern.
This result indicates that the whole-image reconstruction induced by the Stable Diffusion inpainting pipeline accounts for the loss in payload-bearing signal. When the pixels outside the edit mask are restored, payload recovery improves accordingly.

\para{Higher edge density is associated with weaker recovery.} We next ask which characteristics of the watermark region outside the edit mask are associated with the magnitude of the effect.
One hypothesis is that regions containing high-frequency content undergo larger pixel-level value changes from the generative pipeline than smooth, low-frequency regions, such as sky and meadows.
We test this hypothesis by measuring {\it Canny edge density}~\cite{canny1986computational} over watermark-region pixels outside the edit mask, investigating whether it affects payload recovery (Panel C of Figure~\ref{fig:local-edit-log10p}).
Interestingly, for MaskWM, the evidence score decreases monotonically from \(1.53\) in the smoothest quintile to \(0.36\) in the highest-edge quintile. Bit accuracy falls from \(0.560\) to \(0.499\) over the same range.
The same trend also applies to WAM, though less significantly.

This result suggests that payload recovery is inversely correlated with the edge density of the preserved watermark region, indicating that watermarks embedded in high-frequency content may be more susceptible to changes introduced by the evaluated generative pipeline.
This can be concerning for watermark applications because high-frequency regions are often considered favorable for watermark embedding because they allow larger perturbations while preserving imperceptibility~\cite{cox1997secure,hartung1999multimedia}.

\subsection{Synchronization is selective}
\label{sec:syncseal-results}
We also examine synchronization and its effect on payload recovery under transformations, particularly those that change image alignment, since synchronization is designed to correct for exactly this kind of misalignment.
We examine where it helps, where it reduces payload recovery, and what fidelity cost it adds. We adopt SyncSeal as a representative synchronization technique, with extension to other approaches left for future investigation.

\begin{figure}[htbp]
  \centering
  \includegraphics[width=0.85\columnwidth]{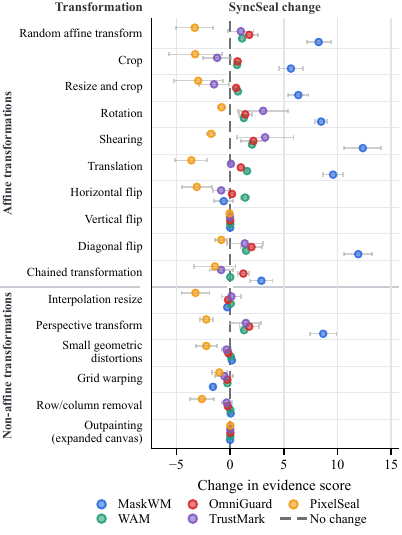}
  \Description{Changes in payload evidence caused by adding SyncSeal under individual alignment transformations, with variation across datasets.}
  \tightcaption{SyncSeal effect on payload recovery under changes in alignment}{Rows include only transformations in the alignment category, grouped into affine and non-affine blocks.
  Points report the change in payload-length-aware evidence score, $-\log_{10}(p_{\mathrm{random}})$, from enabling SyncSeal; positive values indicate stronger payload recovery with SyncSeal.
  Horizontal bars show the sample standard deviation across the four dataset-level changes.}
  \label{fig:sync-align-subattacks}
\end{figure}

\para{SyncSeal helps when transformations match its scope.}
Our results show that SyncSeal's gains concentrate on global geometric mappings within its reported scope in SyncSeal's original paper, consistent with its intended role of estimating and inverting such transformations.
Figure~\ref{fig:sync-align-subattacks} shows how enabling SyncSeal changes the evidence score for the evaluated transformations that specifically alter alignment.
For MaskWM, SyncSeal increases the evidence score by \(+12.36\) on shearing, \(+11.93\) on diagonal flip, \(+9.59\) on translation, \(+8.67\) on perspective transform, \(+8.47\) on rotation, and \(+8.24\) on random affine transform.
However, the gains do not extend to \emph{every} transformation in this category.
Under vertical flip, for example, the evidence score remains close to zero after SyncSeal.

\para{Synchronization can conflict with payload decoding.}
For PixelSeal, the added synchronization signal reduces rather than improves recovery in several cases.
As shown in Fig.~\ref{fig:headline_payload_recovery}, SyncSeal lowers the evidence score for PixelSeal in both the clean setting and the alignment category. The evidence score also drops under signal distortions and indirect local edits when evaluated with SyncSeal.
One possible explanation is that the additional full-image signal shifts the residual seen by the PixelSeal decoder away from its expected input distribution.
The same trend also applies to TrustMark for certain transformations such as cropping, though relatively mildly.

\section{Discussion and Conclusion}
Local watermarking has been proposed to track provenance at the level of individual objects within an image, yet current evaluation of local watermarking methods remains fragmented. We address this gap by developing a comprehensive benchmark spanning five watermarking methods, four datasets, and 55 non-adaptive image transformations, and by analyzing the robustness vulnerabilities of state-of-the-art local watermarking schemes. We now conclude with several key takeaways from our results.

\para{Current methods fail on smaller regions.} Tracing a small object within an image is one of the main motivations for local watermarking in composited imagery, with applications such as e-commerce product image compositing. However, we find that even in the clean setting with no transformations applied, current local watermarking methods perform poorly on smaller region sizes. Future designs of local watermarking methods should improve this capability for smaller regions, for example, by dynamically scaling the number of embedded payload bits according to region area, or by embedding watermark signals across multiple spatial resolutions so that decoding remains robust even when the target region is small.

\para{Payload recovery trades off with image fidelity.} We also find a significant trade-off between payload recovery and image fidelity. For example, MaskWM achieves the strongest clean payload recovery but also produces the lowest clean image fidelity. SyncSeal improves recovery for several transformations but also adds a further fidelity cost across every evaluated method. This trade-off creates a deployment problem. There may be use cases where image fidelity is less critical, such as social media previews, where the watermark's provenance signal matters more than visual quality, but for quality-critical applications such as professional photography or medical imaging, it is unacceptable for image quality to degrade noticeably. Future work should vary the strength of the added signal within each method to identify the point at which recovery gains stop justifying further image distortion.

\para{Generative editing can break a local watermark without even targeting it.}
Inpainting edits affect pixels outside the edit region, which can reduce payload recovery without visible image quality loss in the watermarked region itself. The two outpainting transformations also bring payload recovery down to random guessing. This makes generative edits a serious threat to local watermarking robustness, because modern image processing workflows increasingly apply those edits. We therefore hope future benchmarks will include these types of transformations, and encourage future local watermarking methods to be robust to them.

\para{Robustness vulnerabilities expose a security gap.}
Several transformations in our benchmark reduce payload recovery substantially, sometimes to the level of random guessing. When applied deliberately, these transformations become non-adaptive black-box removal attacks.
Therefore, our results generalize to a broader security concern: if a non-adaptive black-box attacker can already achieve this level of degradation, an adaptive, strategic attacker could in principle perform even better. Future work should evaluate adaptive attacks that may use knowledge of the method design or access to decoder outputs.

\para{Conclusion.}
Local watermarking becomes increasingly important as AI-assisted editing makes it easier to blend human-created and AI generated content within a single image. However, our benchmark shows that, within the evaluated scope, none of the current local watermarking methods is ready for reliable production deployment. Payload recovery performs unsatisfactorily on small target regions, trades off against image fidelity, and remains vulnerable to various transformations, including generative edits. Future work will need to address these challenges before local watermarking can be deployed reliably.
\bibliographystyle{ACM-Reference-Format}
\bibliography{references}

\onecolumn
\appendix

\section{Additional Evaluation Details}\label{sec:additional-evaluation-details}
\begingroup
\small
\noindent
\begin{minipage}[t]{0.49\textwidth}
  \vspace{0pt}
  \subsection{Methodology details}
  \label{sec:empirical-details}

  \para{Images.} For each method, we evaluate a fixed set of 200 images from each dataset, giving 800 images per method. Before embedding, input images are resized to \(512\times512\) pixels, and watermark masks are generated in the same coordinate frame. 

  \para{Masks.} For COCO and SA-1B, we sample uniformly among segmentation masks covering at least \(2\%\) of the image. DIV2K and MIRFLICKR lack segmentation masks, so we generate content-independent masks. For methods that accept arbitrary shapes, rectangles and irregular masks are equally likely. We sample a target area of \(4\%\)--\(14\%\), a log-uniform bounding-box aspect ratio of \(0.6\)--\(1.8\), and a random valid location; irregular masks fill \(55\%\)--\(80\%\) of the box. Because TrustMark requires a rectangle, we use rectangular masks for DIV2K and MIRFLICKR and the smallest rectangle enclosing the selected segmentation mask for COCO and SA-1B.
  
  \para{Transformations.} For each stochastic transformation, such as random cropping, Gaussian noise, or blackout applied to a randomly sampled region, we sample a single realization per image using a fixed random seed. 

  \para{Reproducibility.} We use the released pretrained weights for all watermarking methods and SyncSeal without further training, and fix random seeds for transformation parameters, watermark mask sampling, and payload generation. The benchmark code will be available at \url{https://github.com/garandorinc/markbench}.

  \para{Reporting convention.} Unless otherwise stated, the text reports results without SyncSeal; SyncSeal results and direct comparisons are labeled explicitly. We first average within each dataset and then report the unweighted mean across the four dataset means.
  \subsection{Watermark-Region Area Distributions}
  \appendixnote{Figure~\ref{fig:mask-area-dist} reports the area distributions used in Figure~\ref{fig:watermark-area-vs-perf}.}
  \centering
  \includegraphics[width=0.76\linewidth]{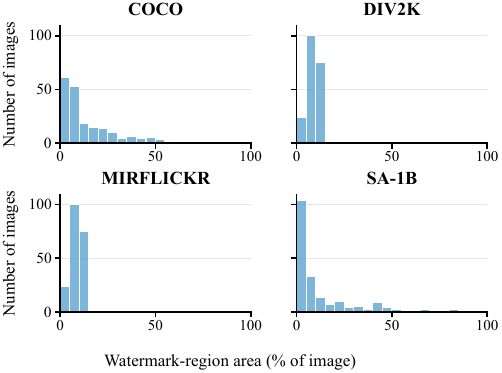}
  \Description{Distributions of watermark-region area across COCO, SA-1B, DIV2K, and MIRFLICKR, with the dataset mean areas indicated.}
  \begingroup
  \captionof{figure}{\textbf{Watermark-region area by dataset.} Distributions for clean MaskWM examples without SyncSeal.
  COCO and SA-1B use native masks; DIV2K and MIRFLICKR use sampled masks.
  Mean areas are 15.22\% (COCO), 12.50\% (SA-1B), and 8.80\% (DIV2K and MIRFLICKR).}
  \label{fig:mask-area-dist}
  \endgroup
\end{minipage}
\hfill
\begin{minipage}[t]{0.49\textwidth}
  \vspace{0pt}
  \subsection{Transformation Protocol Examples}
  \appendixnote{Figure~\ref{fig:protocol-examples} shows the local-edit controls.}
  \begingroup
  \centering
  \includegraphics[width=0.76\linewidth]{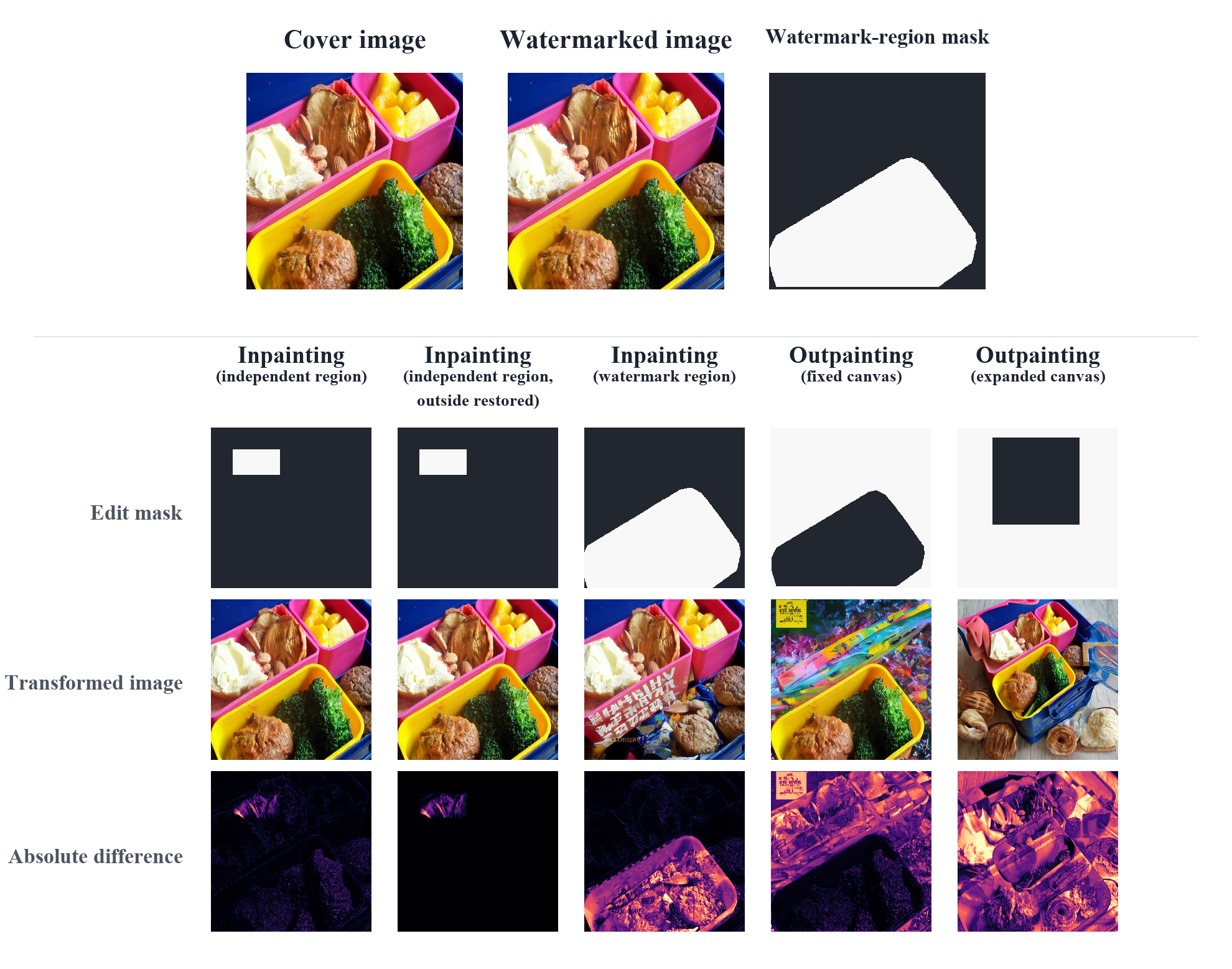}
  \Description{Examples of inpainting and outpainting show the edit mask, transformed image, and absolute pixel difference. The outside-restored control preserves pixels outside the edit mask.}
  \begingroup
  \captionof{figure}{\textbf{Inpainting and outpainting protocols.} Examples show edit masks, transformed images, and absolute differences.
  The outside-restored control restores pixels outside the edit mask; fixed-canvas outpainting regenerates surrounding content, whereas expanded-canvas outpainting extends and rescales the canvas.}
  \label{fig:protocol-examples}
  \endgroup
  \endgroup
  \subsection{Transformation Taxonomy}
  \begingroup
  \centering
\captionof{table}{Taxonomy of the 55 transformations; the clean setting is listed separately.}
\label{tab:attack-taxonomy}
\scriptsize
\setlength{\tabcolsep}{2pt}
\renewcommand{\arraystretch}{0.85}
\begin{tabularx}{\linewidth}{@{}>{\raggedright\arraybackslash}p{0.18\linewidth}>{\raggedright\arraybackslash}p{0.31\linewidth}Y@{}}
\toprule
\textbf{Category} & \textbf{Subcategory} & \textbf{Evaluation conditions} \\
\midrule
Clean & Reference & - \\
\midrule
Signal & Color and tone transformations & brightness, color jitter, randomized color jitter, contrast, gamma correction, grayscale conversion, histogram equalization, color inversion, RGB channel shift, saturation \\
 & Filtering and enhancement & box blur, convolutional filter, Gaussian blur, high-pass filter, low-pass filter, median filter, motion blur, sharpening \\
 & Noise & dither noise, Gaussian noise, Poisson noise, salt-and-pepper noise, speckle noise \\
 & Compression and quantization & color quantization, GIF compression, JPEG compression, JPEG 2000 compression, posterization, uniform quantization \\
 & Frequency-domain transformations & DCT transform, DWT transform, FFT transform \\
\midrule
Alignment & Geometric transformations & random affine transform, crop, interpolation resize, perspective transform, resize and crop, rotation, shearing, small geometric distortions, translation, grid warping \\
 & Diagonal flip & diagonal flip \\
 & Horizontal flip & horizontal flip \\
 & Vertical flip & vertical flip \\
 & Row/column removal & row/column removal \\
 & Chained transformation & JPEG + crop + brightness chain \\
 & Outpainting (expanded canvas) & outpainting (expanded canvas) \\
\midrule
Indirect local edits & Blackout (independent region) & blackout (independent region) \\
 & Inpainting (independent region) & inpainting (independent region) \\
 & Inpainting (independent region, outside restored) & inpainting (independent region, outside restored) \\
 & Splicing (independent region) & splicing (independent region) \\
 & Outpainting (fixed canvas) & outpainting (fixed canvas) \\
\midrule
Direct watermark edits & Inpainting (watermark region) & inpainting (watermark region) \\
 & Splicing (watermark region) & splicing (watermark region) \\
\bottomrule
\end{tabularx}

  \endgroup
\end{minipage}
\endgroup

\clearpage
\subsection{Local Adaptation of Global Watermarking Methods}
\label{app:local-adaptations}

This section describes how we adapt three global watermarking methods, OmniGuard, TrustMark, and PixelSeal, to the local watermarking setting evaluated in this work.
For each method, we explain how we restrict payload embedding to a selected image region and how the payload is subsequently recovered during decoding.
We use the released pretrained weights without additional training or fine-tuning.
During embedding, a binary mask specifies the selected region.
We refer to the pixelwise difference between an image before and after embedding as the \emph{watermark residual}: it records the pixel changes introduced by the embedder.

\paragraph{OmniGuard.}
OmniGuard embeds two distinct signals: an auxiliary signal for tamper localization and a binary payload for copyright identification.
We first apply its auxiliary embedding stage across the entire image.
We then resize this intermediate image to \(256\times256\) and pass it through the binary payload encoder.
Subtracting the encoder's input from its output gives the changes introduced specifically by payload embedding.
We resize this payload residual back to the benchmark image resolution of \(512\times512\), retain it only within the selected mask, and add it to the intermediate image.
Consequently, the auxiliary signal remains across the image, while the binary payload signal is confined to the selected region.
For payload recovery, we resize the complete received image to \(256\times256\) and apply the payload decoder.

\paragraph{TrustMark.}
TrustMark operates on rectangular regions.
We therefore replace the selected mask with its smallest enclosing axis-aligned bounding box, crop that region, and resize the crop to the encoder's native resolution.
We compute the watermark residual from the encoder's output and subtract its spatial mean separately for each color channel.
We then resize the residual back to the crop's original dimensions and add it to the corresponding region in the original image.
During decoding, TrustMark's region detector first predicts bounding boxes from the received image.
The predicted regions are cropped and resized to the decoder's native resolution for payload extraction.
We use the Q variant with a 100-bit payload and error correction disabled.

\paragraph{PixelSeal.}
We first apply PixelSeal's embedder to the complete image to obtain a globally watermarked output.
We subtract the original image from this output, multiply the resulting residual by the selected mask, and add the masked residual back to the original image.
This confines all changes introduced by PixelSeal to the selected region.
Unlike OmniGuard, this adaptation does not retain a separate auxiliary watermark outside that region.
We use the released 256-bit model and decode the payload from the complete received image using its whole-image decoder.

For all three methods, the selected mask is used only during embedding.
Decoding receives neither the ground-truth mask nor its bounding box.

\subsection{Transformation Parameters}
\label{app:transformation-parameters}

Table~\ref{tab:attack-implementation-summary} describes the clean reference and all 55 transformation conditions.
For each image and transformation condition, we sample one predefined parameter configuration using fixed random seeds.
For brightness and Gaussian blur, intervals denote uniform sampling within the selected configuration; the listed values are not an exhaustive sweep applied to every image.
In the Gaussian blur configurations, \(k\) denotes a square \(k\times k\) kernel unless a rectangular kernel is specified, and \(U[a,b]\) denotes a uniform draw from \([a,b]\).

{\small
\setlength{\LTleft}{0pt}
\setlength{\LTright}{\fill}
\setlength{\LTcapwidth}{\textwidth}
\setlength{\tabcolsep}{2pt}
\renewcommand{\arraystretch}{1.08}
\begin{longtable}{@{}>{\raggedright\arraybackslash}p{0.12\textwidth} >{\raggedright\arraybackslash}p{0.21\textwidth} >{\raggedright\arraybackslash}p{\dimexpr0.67\textwidth-4\tabcolsep\relax}@{}}
\caption{Definitions of the clean reference and 55 transformation conditions. The clean reference is listed separately; each transformation row gives its category, implementation, and parameter settings.}
\label{tab:attack-implementation-summary}\\
\toprule
\textbf{Category} & \textbf{Condition} & \textbf{Definition} \\
\midrule
\endfirsthead
\multicolumn{3}{@{}l}{\footnotesize\textit{Table~\thetable\ continued.}}\\
\addlinespace[1pt]
\toprule
\textbf{Category} & \textbf{Condition} & \textbf{Definition} \\
\midrule
\endhead
\midrule
\multicolumn{3}{r}{Continued on next page}\\
\endfoot
\bottomrule
\endlastfoot
Clean & Clean reference & No transformation is applied; the watermarked image is decoded directly. \\
\addlinespace[2pt]
Alignment & Random affine transform & Randomly samples affine parameters: rotation up to 180 degrees, x/y scale between 0.6 and 1.4, shear up to 0.3, and translation up to 0.2 in normalized coordinates. Ground-truth localization masks are transformed with the same operation. \\
Alignment & Crop & Keeps a randomly located square crop with side ratio 0.32-0.95 and discards the rest of the frame. Ground-truth localization masks are cropped in the same way. \\
Alignment & Resize and crop & Samples a crop area ratio from 0.10-1.0 and aspect ratio from 0.3-3.0, then resizes the crop back to the original image size. Ground-truth localization masks follow the same crop-and-resize transform. \\
Alignment & Interpolation resize & Resizes the image with linear interpolation using fixed scale factors of 0.5, 0.75, 1.25, or 1.5, or a random factor between 0.5 and 1.5. Ground-truth localization masks use the same resizing factors. \\
Alignment & Perspective transform & Applies a projective corner perturbation with distortion scale 0.1-0.8. The image and ground-truth localization masks share the same perspective map. \\
Alignment & Rotation & Rotates by sampled ranges of +/-10, +/-30, or +/-90 degrees, or by fixed angles of 5, 10, 30, 45, 60, or 90 degrees; ground-truth localization masks use the same rotation. \\
Alignment & Shearing & Applies horizontal and vertical shear factors of 0.1, 0.2, or 0.3. Ground-truth localization masks are transformed with the same operation. \\
Alignment & Small geometric distortions & Applies a mild nonrigid warp from small corner shifts, sinusoidal bending, and local wave perturbations. Ground-truth localization masks are resampled with the same grid. \\
Alignment & Translation & Shifts the image by 5, 10, or 20 pixels in both axes. Ground-truth localization masks are shifted consistently. \\
Alignment & Grid warping & Applies a sinusoidal grid warp with one-pixel default amplitude and grid period 20 pixels. Ground-truth localization masks are warped with the same sampling grid. \\
Alignment & Horizontal flip & Mirrors the image left to right and applies the same coordinate change to ground-truth localization masks. \\
Alignment & Vertical flip & Mirrors the image top to bottom and applies the same coordinate change to ground-truth localization masks. \\
Alignment & Diagonal flip & Transposes the image across its diagonal, swapping the horizontal and vertical axes of the image and ground-truth localization masks. \\
Alignment & Row/column removal & Randomly removes 10, 30, or 50 rows and the same number of columns, producing a smaller image frame. Ground-truth localization masks use the same retained indices. \\
Alignment & JPEG + crop + brightness chain & Applies JPEG compression at quality 40, 60, or 80, keeps a random 0.71-side crop, and then multiplies intensity by 0.5. \\
Alignment & Outpainting (expanded canvas) & Places the full watermarked image on a larger canvas, scaled by 1.5-2.0, optionally rescales the pasted image by 0.8-1.2, and uses inpainting to synthesize the surrounding canvas. \\
\addlinespace[2pt]
Signal & Brightness & Multiplies image intensity by a factor selected from the fixed values \(\{0.1,0.25,0.5,0.75,1,1.25,1.5,1.75,2\}\) or the uniform distributions \(U[0.5,1.5]\) and \(U[0.7,1.3]\); clips the result to the valid pixel range. \\
Signal & Contrast & Changes contrast around the image mean with factors from 0.5 to 2.0, including sampled ranges 0.5-1.5, 0.5-2.0, and 0.7-1.3. \\
Signal & Saturation & Changes color saturation with factors 0.1, 0.5, sampled 0.5-2.0, sampled 0.7-1.3, 1.0, or 1.5 while preserving image geometry. \\
Signal & Gamma correction & Applies nonlinear intensity remapping with gamma values 0.5, 0.8, 1.6, and 2.5. \\
Signal & Color jitter & Shifts hue by \(0.1,0.2,-0.1,\) or \(-0.2\) radians, or by a value sampled uniformly from \([-0.1,0.1]\), while leaving brightness, contrast, and saturation unchanged. \\
Signal & Randomized color jitter & Applies Kornia ColorJiggle using one of three amplitude tuples \((b,c,s,h)\): \((0.05,0.05,0.05,0.01)\), \((0.1,0.1,0.1,0.02)\), or \((0.1,0.1,0.1,0.05)\), where \(b,c,s,h\) specify brightness, contrast, saturation, and hue jitter. \\
Signal & RGB channel shift & Adds symmetric per-channel RGB shifts with limits 0.02, 0.05, or 0.1 in normalized color units. \\
Signal & Grayscale conversion & Converts the image to grayscale and keeps it in a three-channel representation for decoding. \\
Signal & Histogram equalization & Redistributes image intensities to equalize the histogram while keeping the image frame fixed. \\
Signal & Color inversion & Replaces each color value by its complement. \\
Signal & Box blur & Applies a mean filter with kernel size 3, 5, or 7 pixels. \\
Signal & Gaussian blur & Samples one kernel/standard-deviation configuration \((k,\sigma)\): \((3,0.5)\), \((3,U[0.1,1])\), \((3,3)\), \((5,1)\), \((5,U[0.1,1.5])\), \((7,U[0.1,2])\), \((9,1.5)\), \((13,2)\), \((17,2)\), or \((3,5)\); an additional configuration uses a rectangular \(3\times17\) kernel with \(\sigma\sim U[0.1,2]\). \\
Signal & Motion blur & Applies directional blur kernels of size 3, 5, 7, or 9 to simulate camera or object motion. \\
Signal & Sharpening & Enhances local contrast with sharpening factors 0.5, 1.0, or 2.5. \\
Signal & High-pass filter & Suppresses low-frequency image content and emphasizes edges and fine texture. \\
Signal & Low-pass filter & Suppresses high-frequency image content with kernel sizes 3, 5, or 7, keeping smoother low-frequency structure. \\
Signal & Convolutional filter & Applies a fixed 3-by-3 edge-enhancing sharpening kernel. \\
Signal & Median filter & Replaces each pixel by the median of a 3-, 5-, or 7-pixel local neighborhood, reducing impulse-like noise. \\
Signal & Gaussian noise & Adds zero-mean Gaussian noise with standard deviation from 1.0 to 25.5 on the 0-255 pixel scale. \\
Signal & Poisson noise & Applies signal-dependent shot noise with scale factors 0.75, 1.0, or 1.25. \\
Signal & Speckle noise & Adds multiplicative noise proportional to local image intensity with standard deviation 0.05 or 0.15. \\
Signal & Dither noise & Applies dithering noise with intensity 0.02, 0.08, or 0.12. \\
Signal & Salt-and-pepper noise & Randomly sets 5\% or 10\% of pixels to black or white. \\
Signal & JPEG compression & Encodes and decodes the image with JPEG quality 40, 50, 60, 70, 80, 90, or 95. \\
Signal & JPEG 2000 compression & Encodes and decodes the image with JPEG 2000 compression ratios 35 or 100. \\
Signal & GIF compression & Converts the image through a palette-based GIF representation. \\
Signal & Posterization & Keeps 3, 4, 5, or 6 bits per color channel by dropping lower-order bits. \\
Signal & Uniform quantization & Maps pixel values onto 8, 16, or 32 uniformly spaced intensity levels. \\
Signal & Color quantization & Reduces the color palette to 256, 128, 64, 32, or 16 representative colors. \\
Signal & DCT transform & Perturbs discrete cosine transform coefficients with strength levels 1-10 and power 0.01-0.1, with optional coefficient shuffling. \\
Signal & DWT transform & Perturbs Haar or Daubechies wavelet coefficients with strength levels 1-15 and power 0.05-0.5. \\
Signal & FFT transform & Perturbs Fourier-domain coefficients with strength levels 1-15 and power 0.001-0.05, with optional coefficient shuffling. \\
\addlinespace[2pt]
Indirect local edits & Blackout (independent region) & Places a black square occluder with side length 64, 128, or 256 pixels at a randomly sampled image location. The edit region is chosen independently of the watermark region. \\
Indirect local edits & Inpainting (independent region) & Samples a rectangular edit region independently of the watermark region, with side ratios between 0.05 and one of 0.15, 0.35, or 0.50, then replaces that region with an inpainted fill. \\
Indirect local edits & Inpainting (independent region, outside restored) & Uses the same rectangular-mask sampling protocol as inpainting (independent region), with independently sampled edit regions, and restores all pixels outside the edit region from the watermarked input after generation. \\
Indirect local edits & Splicing (independent region) & Pastes donor content from another image in the same dataset into one or more independently sampled edit regions. \\
Indirect local edits & Outpainting (fixed canvas) & Uses the complement of the watermark region as the generation mask at the original image size, without restoring unmasked pixels after generation. \\
\addlinespace[2pt]
Direct watermark edits & Inpainting (watermark region) & Uses the watermark region dilated with an odd square kernel whose side length is approximately 3\% of the shorter image side (at least three pixels) as the edit region, and replaces it with an inpainted fill. \\
Direct watermark edits & Splicing (watermark region) & Pastes donor content from another image in the same dataset into the watermark region. \\
\end{longtable}
}

\section*{Use of Generative AI}
We used large language models only to assist with grammar checking, language polishing, and coding assistance.
The authors reviewed all suggested edits and remain responsible for the final manuscript.

\end{document}